\documentclass[a4paper]{eccomas_paper-2024}

\usepackage{graphicx}
\usepackage{amsmath,amsfonts,amssymb}
\usepackage{bm}
\usepackage{booktabs}
\usepackage[ plainpages = false, pdfpagelabels,
	     bookmarks,
	     bookmarksopen = true,
	     bookmarksnumbered = true,
	     breaklinks = true,
	     linktocpage,
	     pagebackref,         
	     colorlinks = true,  
         hypertexnames=false,   
	     linkcolor = green!50!black,
	     urlcolor  = blue!50!black,
	     citecolor = blue!50!black,
	     anchorcolor = blue,
	     hyperindex = true,
	     hyperfigures
	     ]{hyperref}
\usepackage[capitalise,noabbrev]{cleveref}
\usepackage{algorithm}
\usepackage{algorithmicx}
\usepackage{algpseudocode}
\usepackage{comment}
\usepackage{tikz}
\newlength{\figureheight}
\newlength{\figurewidth}
\usepackage{tikzsymbols}
\usepackage{pgfplots}
\pgfplotsset{compat=newest} 
\usetikzlibrary{plotmarks}
\usetikzlibrary{arrows.meta}
\usetikzlibrary{positioning}
\usepgfplotslibrary{patchplots}
\usepackage{grffile}
\usepackage{multirow}
\usepackage{pdflscape}
 \usepackage{fontawesome}
\pgfplotsset{every axis/.append style={
                    label style={font=\scriptsize},
                    tick label style={font=\scriptsize},
                    legend style={font=\scriptsize}
                    }}
\pgfplotsset{compat=newest}
\pgfplotsset{plot coordinates/math parser=false}
\pgfplotsset{grid style={dotted,gray}}
\usepgfplotslibrary{groupplots}
\pgfdeclarelayer{background}
\pgfdeclarelayer{foreground}
\pgfsetlayers{background,main,foreground}
\usepackage{soul}
\usepackage{stmaryrd}
\usepackage{subcaption}
\usepackage{textcomp}
\usepackage{xcolor}

\makeatletter
\newcommand{\opnorm}{\@ifstar\@opnorms\@opnorm}
\newcommand{\@opnorms}[1]{%
  \left|\mkern-1.5mu\left|\mkern-1.5mu\left|
   #1
  \right|\mkern-1.5mu\right|\mkern-1.5mu\right|
}
\newcommand{\@opnorm}[2][]{%
  \mathopen{#1|\mkern-1.5mu#1|\mkern-1.5mu#1|}
  #2
  \mathclose{#1|\mkern-1.5mu#1|\mkern-1.5mu#1|}
}
\makeatother

\newcommand{\norm}[1]{\left\lVert#1\right\rVert}

\newcommand{\nint}[1]{\llbracket#1\rrbracket}

\def\defequal{\stackrel{\mbox{\footnotesize def}}{=}}

\def\*#1{\mathbf{#1}}

\newcommand{\nucnorm}[1]{\norm{#1}_{*}}
\newcommand{\fnorm}[1]{\norm{#1}_{\mathrm{F}}}
\newcommand{\spnorm}[1]{\norm{#1}_{2}}

\newcommand{\bSigma}{\boldsymbol{\Sigma}}

\newcommand{\mR}{\mathcal{R}}
\newcommand{\mT}{\mathcal{T}}
\newcommand{\mU}{\mathcal{U}}

\newcommand{\RR}{\mathbb{R}}

\newcommand{\minimize}[2]{\ensuremath{\underset{\substack{{#1}}}%
{\mathrm{minimize}}\;\;#2 }}

\DeclareMathOperator{\Loss}{\mathcal{L}}

\SetSymbolFont{stmry}{bold}{U}{stmry}{m}{n}

\newcommand{\define}[1]{\textit{#1}}

\graphicspath{{img/}}

\title{Automated transport separation using the neural shifted proper orthogonal
  decomposition.}

\author{Beata Zorawski$^1$, Shubhaditya Burela$^1$, Philipp Krah$^2$, Arthur Marmin$^2$ and Kai Schneider$^2$}

\heading{Beata Zorawski, Shubhaditya Burela, Philipp Krah, Arthur Marmin, and Kai Schneider}

\address{$^{1}$ Institute of Mathematics\\
  Technische Universität Berlin\\
  Str. des 17. Juni 136, 10587 Berlin, Germany\\
  e-mail: beatazorawski@gmail.com, burela@tnt.tu-berln.de
\and
$^{2}$ Aix-Marseille Université, CNRS, I2M, UMR 7373\\
3 place Victor Hugo, 13003 Marseille, France\\
e-mail: philipp.krah@univ-amu.fr, arthur.marmin@univ-amu.fr, kai.schneider@univ-amu.fr
}


\abstract{
This paper presents a neural network-based methodology for the decomposition of transport-dominated fields using the shifted proper orthogonal decomposition (sPOD).
Classical sPOD methods typically require an a priori knowledge of the transport operators to determine the co-moving fields.
However, in many real-life problems, such knowledge is difficult or even impossible to obtain, limiting the applicability and benefits of the sPOD.
To address this issue, our approach estimates both the transport and co-moving fields simultaneously using neural networks.
This is achieved by training two sub-networks dedicated to learning the transports and the co-moving fields, respectively.
Applications to synthetic data and a wildland fire model illustrate the capabilities and efficiency of this neural sPOD approach, demonstrating its ability to separate the different fields effectively.
  
}

\begin{document}
\thispagestyle{empty}


\section{Introduction}

This paper focuses on low-rank decompositions for transport-dominated fields, as they occur in \define{model order reduction} (MOR) of wildland fires \cite{Black_F_2021_j-fluids_efficient_wfsnmor,
  Burela_S_2023_PP_parametrc_morwfmspbdlm} or other technical fluid applications \cite{Huang_C_2018_p-joint_prop-conf_challenges_romrf,
  Kovarnova_A_2023_p-topical-pb-fluid-mech_model_orplfsrdto,
  Nonino_M_2023_j-adv-comput-sci-eng_reduced_bmmtmfsipsdkw}.
Our research aims to derive low-dimensional surrogate models that incorporate the physical information of the transport to enable more efficient flow simulations.
Furthermore, we are interested in separating flows with multiple transports to study the individual transport systems and their interaction independently.

An efficient method that separates transports is the \define{shifted proper orthogonal decomposition} (sPOD) \cite{Reiss_J_2018_j-siam-j-sci-comput_shifted_podmdmtp}.
Compared to other low-rank decomposition methods for transports~\cite{Barral_N_2024_j-comput-phys_registration_mrppdesa,
  Krah_P_2023_j-sci-comput_front_trcmfnmrardekpprt,
  Mojgani_R_2023_j-comput-meth-appl-mech-eng_kolmogorov_wlpinnccmcdpde,
  Rim_D_2023_j-siam-j-sci-comput_manifold_atsmrtdp,
  Taddei_T_2020_j-siam-j-sci-comput_registration_mmordcgr}, the sPOD has the unique feature of giving direct access to the transported quantities and enabling a strict separation of the corresponding fields.
This method requires an apriori knowledge of the transport $\{\Delta^{k}\}_{k}$.
With this information, the data $q(x,t)$ can be represented with several low-rank co-moving fields $\{q^{k}\}_{k}$
\begin{equation*}
  q(x,t) =\sum_{k=1}^{K} q^k(x-\Delta^k(t),t) \, .
\end{equation*}
Knowing $\{\Delta^k\}_k$, the individual fields $\{q^{k}\}_{k}$ can be determined with an optimization procedure~\cite{Krah_P_2024_PP_robust_spodpmdfmt,
  Reiss_J_2021_j-siam-j-sci-comput_optimization_mdsmt,
  Reiss_J_2018_j-siam-j-sci-comput_shifted_podmdmtp}.
The $\{q^{k}\}_{k}$ describe each traveling wave in their co-moving frame.
The dynamics of $q^k$ are assumed to be slow so that it can be decomposed in a low-rank fashion. 
In this work, we extend the approach to automatically detect the shift-transformations $\{\Delta^k\}_k$ and determine the fields $\{q^{k}\}_{k}$ using neural networks.

Similar approaches that estimate both the transformations and co-moving fields simultaneously can be found in the literature~\cite{Barral_N_2024_j-comput-phys_registration_mrppdesa,
  Black_F_2021_book_modal_dfdgto,
  Krah_P_2023_j-sci-comput_front_trcmfnmrardekpprt,
  Mendible_A_2020_j-theo-comput-fluid-dynam_dimensionality_rromtwp,
  Mojgani_R_2023_j-comput-meth-appl-mech-eng_kolmogorov_wlpinnccmcdpde, 
  Papapicco_D_2022_j-comput-meth-appl-mech-eng_neural_nspodmlanrhe,
  Rim_D_2023_j-siam-j-sci-comput_manifold_atsmrtdp,
  Taddei_T_2020_j-siam-j-sci-comput_registration_mmordcgr}.
However, most of them rely on one-to-one mappings that only allow single transported waves \cite{Barral_N_2024_j-comput-phys_registration_mrppdesa,
  Mojgani_R_2023_j-comput-meth-appl-mech-eng_kolmogorov_wlpinnccmcdpde,
  Papapicco_D_2022_j-comput-meth-appl-mech-eng_neural_nspodmlanrhe,
  Rim_D_2023_j-siam-j-sci-comput_manifold_atsmrtdp}.
The mappings are either found using neural networks~\cite{Mojgani_R_2023_j-comput-meth-appl-mech-eng_kolmogorov_wlpinnccmcdpde,
  Papapicco_D_2022_j-comput-meth-appl-mech-eng_neural_nspodmlanrhe}, assuming the existence of a characteristic map~\cite{Rim_D_2023_j-siam-j-sci-comput_manifold_atsmrtdp} or a structure registration~\cite{Barral_N_2024_j-comput-phys_registration_mrppdesa}.
In contrast, \cite{Krah_P_2023_j-sci-comput_front_trcmfnmrardekpprt} allow more complex mappings that also include topological changes for splitting and merging reaction waves.

When multiple waves are traveling, the sPOD method is particularly advantageous.
However, the quality of the decomposition critically depends on identifying suitable shift transformations.
Consequently, \cite{Black_F_2021_book_modal_dfdgto,
  Mendible_A_2020_j-theo-comput-fluid-dynam_dimensionality_rromtwp} also attempted to optimize the shifts.
In~\cite{Mendible_A_2020_j-theo-comput-fluid-dynam_dimensionality_rromtwp}, a two-phase optimization is proposed: first, the shifts $\{\Delta^{k}\}_{k}$ of traveling waves are fitted using a dictionary learning approach, followed by the determination of the co-moving fields.
In contrast, \cite{Black_F_2021_book_modal_dfdgto} optimizes both simultaneously.
Although \cite{Mendible_A_2020_j-theo-comput-fluid-dynam_dimensionality_rromtwp} appears less dependent on the initial guess of the paths, optimizing $\{\Delta^k\}_k$ and $\{q^{k}\}_k$ separately may be sub-optimal.
Unfortunately, in~\cite{Black_F_2021_book_modal_dfdgto}, the initial path need to be already close to the optimum in order to provide correct results.

In this work, we are moving away from a two-phase optimization and instead, we leverage the expressivity of neural networks to identify the correct paths.
The advantage over conventional autoencoder model order reduction approaches, such as \cite{Carlberg_K_2013_j-comput-phys_gnat_mnmreiacfdtf}, is that our network structure preserves the translation symmetry of the governing equations, resulting in better interpretability of the resulting decomposition.
This approach shares similarities with~\cite{Papapicco_D_2022_j-comput-meth-appl-mech-eng_neural_nspodmlanrhe}, applying a neural network to deduce the shift operator.
However,~\cite{Papapicco_D_2022_j-comput-meth-appl-mech-eng_neural_nspodmlanrhe} assumes a bijective mapping and only a single transported field, which is always mapped to a reference state in order to promote low-rankness.
Our approach is generally for multiple transported fields and does not require a reference state.

Our paper is organized as follows: Section~\ref{sec:spod} introduces the shifted
proper orthogonal decomposition.
Section~\ref{sec:pb} describes our optimization of interest and its
discretization while our methodology to solve the latter is detailed in
Section~\ref{sec:nspod}.
Numerical results are presented in Section~\ref{sec:simul} and
Section~\ref{sec:conclusion} concludes and draws perspectives.

\paragraph{Notation:}
We use the following notation: matrices are denoted in bold upper case letters,
vectors are denoted in bold lowercase.
We write $\nucnorm{.}$, $\spnorm{.}$, and $\fnorm{.}$ for the nuclear, the
spectral and the Frobenius norms of a matrix respectively.
The set $\nint{1,N}$ denotes the set of natural integers from $1$ to $N$.
For the sake of clarity, we will denote a sequence ${(u_{n})}_{n\in\nint{1,N}}$
using curly bracket $\{u_{n}\}$.


\section{Shifted proper orthogonal decomposition}
\label{sec:spod}

The sPOD performs a non-linear decomposition of a
transport-dominated field $q:\Omega\times\tau\to\RR$ into a sum of multiple co-moving fields $\{q^{k}\}$ that
are transformed by $\{\mT^{k}\}$, where $\Omega$ and $\tau$ are the space and time domains, respectively.
Namely, it aims at searching for $\{q^{k}\}$ and $\{\mT^{k}\}$ such that
\begin{equation}
  \label{eq:sPOD}
  q(x,t) =\sum_{k=1}^{K}\mT^k q^k(x,t) \, ,
\end{equation}
where $K$ is the number of co-moving frames.
In general, the transformations $\{\mT^{k}\}$ are chosen such that the co-moving
fields $\{q^{k}\}$ can be described with the help of a dyadic decomposition
\begin{equation}
  \label{eq:dyadic_decomp}
  q^k(x,t) \approx \sum_{r=1}^{R_{k}} \alpha^k_r(x,t)\phi^k_r(x,t) \, ,
\end{equation}
where $R_{k}$ is the rank of the co-moving field $q^{k}$.
Indeed, if $q$ is the solution of a partial differential equation, then its
representation using~\eqref{eq:sPOD} and~\eqref{eq:dyadic_decomp} results in a
significantly faster \define{reduced order model} (ROM) as shown in \cite{Black_F_2021_j-fluids_efficient_wfsnmor,
  Burela_S_2023_PP_parametrc_morwfmspbdlm}.

In this work, we assume that the transformations $\{\mT^{k}\}$ take the form of
time-dependent shifts $\{\Delta^k\}$ which reads
\begin{equation}
  \label{eq:sPOD-shift}
  \mT^{k} q^{k}(x, t) = q^{k}(x - \Delta^k(t), t) \, .
\end{equation}
In practice, these transformations can be more general but need to be at least
piecewise differentiable in time and diffeomorphic.
For instance, they include rotations~\cite{Kovarnova_A_2023_p-topical-pb-fluid-mech_model_orplfsrdto,
  Kovarnova_A_2022_p-topical-pb-fluid-mech_shifted_podanntcromtds}.
We emphasize that $\{\mT^{k}\}$ are non-linear operators due to their time dependency.
This non-linearity provides great flexibility and expressivity to the sPOD but
also complicates the search for such a decomposition.

Decomposition~\eqref{eq:sPOD} of a transport-dominated field $q$ is usually
obtained with the help of variational methods, where one optimizes a
well-chosen objective function while incorporating the low-rankness
condition~\eqref{eq:dyadic_decomp} on the co-moving fields $\{q^{k}\}$.
Most of the time, a priori knowledge of the transformations $\{\mT^{k}\}$ is necessary
and is obtained from knowledge of the physics phenomena underlying the
application~\cite{Black_F_2021_j-fluids_efficient_wfsnmor,
  Burela_S_2023_PP_parametrc_morwfmspbdlm,
  Kovarnova_A_2022_p-topical-pb-fluid-mech_shifted_podanntcromtds,
  Kovarnova_A_2023_p-topical-pb-fluid-mech_model_orplfsrdto,
  Nicolini_J_2021_j-ieee-trans-plasma-sci_reduced_omadkppspod,
  Reiss_J_2021_j-siam-j-sci-comput_optimization_mdsmt,
  Schulze_P_2018_book_model_rpdcspod}.
However, for a wide range of problems, knowing the shifts a priori is highly
complicated or even impossible.
This is a severe limitation for the applicability of sPOD\@.


\section{Variational formulation and discretization}
\label{sec:pb}
The following states the generalized optimization problem and its discretized formulation.


\paragraph{Continuous problem description}

The sPOD decomposition of a transport-dominated field $q$, can be determined by solving the following optimization problem
\begin{equation}
  \label{eq:main_pb}
  \minimize{\{q^{k}\},\{\mT^{k}\}} \quad
  \norm{q - \sum_{k=1}^{K}\mT^{k}q^{k}} + \lambda \mR(\{q^{k}\})
  \, ,
\end{equation}
where $\norm{.}$ is a functional norm and $\mR$ is a regularization with regularizing parameter $\lambda>0$  that
enforces the low-rankness property of the co-moving fields.

Optimizing~\eqref{eq:main_pb} for given $\{\mT^{k}\}$ has been addressed in several publications already \cite{Krah_P_2024_PP_robust_spodpmdfmt,
  Reiss_J_2021_j-siam-j-sci-comput_optimization_mdsmt,
  Reiss_J_2018_j-siam-j-sci-comput_shifted_podmdmtp}.
In this work, our goal is to solve~\eqref{eq:main_pb} jointly in the co-moving
frames $\{q^{k}\}$ and in the transformations $\{\mT^{k}\}$.
To the best of our knowledge, such a joint optimization approach for the sPOD optimization problem has not been studied except in~\cite{Black_F_2021_book_modal_dfdgto}.
The main reason is the high computational complexity of~\eqref{eq:main_pb} and its
non-convexity.
However, in~\cite{Black_F_2021_book_modal_dfdgto}, the joint optimization is performed with a gradient-based approach, where the initial shifts are presumed to be near the optimal ones.
In contrast, we propose here a black-box modeling strategy relying on neural networks where
the co-moving fields and the transformations are represented by two different sub-networks that
are combined and trained jointly.
In particular, our method  does not require the initial shifts to be close to their
optimal positions.


\paragraph{Discretization}

Numerically, Problem~\cref{eq:main_pb} is solved on a discrete dataset. 
For the sake of clarity and without loss of generality, we assume one spatial
and one temporal dimension which is discretized using $M$ spatial grid points
$\{x_{m}\}$ and $N$ time grid points $\{t_{n}\}$, respectively.
We note $\*Q={(q(x_{m},t_{n}))}_{(m,n)\in\nint{1,M}\times\nint{1,N}}$ the snapshot
matrix with its columns representing the snapshots at each time step. These snapshots are the values of the transport-dominated field $q$ computed on the
uniform discretized grid. 
Similarly, for all $k$ in $\nint{1,K}$,
$\*T^{k}\*Q^{k}={((\mT^{k}q^{k})(x_{m},t_{n}))}_{(m,n)\in\nint{1,M}\times\nint{1,N}}$
is the matrix representing the values of the transformed fields $\mT^{k}q^{k}$ on the
discretized grid, while $\*Q^{k}={(q^{k}(x_{m},t_{n}))}_{(m,n)\in\nint{1,M}\times\nint{1,N}}$
is the matrix representing the values of the co-moving fields in their corresponding frames.

The low-rankness condition on the co-moving fields can now be expressed
using the nuclear norm of the matrices $\{\*Q^k\}$.
Indeed, the nuclear norm is a well-suited surrogate for the rank function since
it is its convex hull.
However, note that relaxing a sum of ranks into a sum of nuclear norms is not
tight since the convex hull of a sum of functions is not equal to the sum of each convex hull in general.
Therefore, the discretized version of
Problem~\eqref{eq:main_pb} can be formulated as
\begin{equation}
  \label{eq:discr_pb}
  \minimize{\{\*Q^{k}\},\{\*T^{k}\}} \quad \Loss(\{\*Q^{k}\},\{\*T^{k}\}) \defequal
  \fnorm{\*Q - \sum_{k=1}^{K}\*T^{k}\*Q^{k}}^{2} + \lambda \sum_{k=1}^{k}\nucnorm{\*Q^{k}} \, .
\end{equation}

Note that, in contrast with previous approaches~\cite{Krah_P_2024_PP_robust_spodpmdfmt,Reiss_J_2021_j-siam-j-sci-comput_optimization_mdsmt,Reiss_J_2018_j-siam-j-sci-comput_shifted_podmdmtp,Black_F_2021_book_modal_dfdgto},
we perform the discretization on the transformed fields instead of transforming
the discretized co-moving fields.
This choice eliminates the need for interpolation when transforming the data on the discretization grid, thus eliminating the associated interpolation error that may be significant.

\section{Neural network approach for joint learning}
\label{sec:nspod}

\begin{figure}[htp!]
  \centering
  \setlength\figureheight{0.43\linewidth}
  \setlength\figurewidth{0.43\linewidth}
  \usetikzlibrary{positioning, shapes.geometric, arrows.meta}
\begin{tikzpicture}[
    node distance=1.5cm and 2cm,
    mynode2/.style={draw, rectangle, minimum width=2cm, minimum height=1cm, align=center},
    arrow/.style={-{Latex[length=3mm, width=2mm]}}
]
\node[mynode2] (inputt) at (0,0) {Input $t$};
\node[mynode2] (inputx) at (0,-2) {Input $x$};

\node[mynode2] (shiftNet) at (6,0) {ShiftNet};
\node[mynode2] (shift) at (12,0) {$\{\Delta^{k}(t)\}$};
\node[mynode2] (x_shifted) at (12,-2) {$\{x+\Delta^{k}(t)\}$};

\node[mynode2] (shapeNet) at (6,-2) {ShapeNet};
\node[mynode2] (outputf) at (3,-4) {Output $\{q^{k}(x,t)\}$};
\node[mynode2] (outputfun) at (9,-4) {Output $\{q^{k}(x+\Delta^{k}(t),t)\}$};

\draw[arrow] (inputt) -- (shiftNet);
\draw[arrow] (inputt) -- (shapeNet);
\draw[arrow] (inputx) -- node[midway, below] {unshifted path} (shapeNet);

\draw[arrow] (shiftNet) -- (shift);

\draw[arrow] (shift) -- (x_shifted);
\draw[arrow] (x_shifted) -- node[midway, below] {shifted path} (shapeNet);
\draw[arrow] (shapeNet) -- (outputf);
\draw[arrow] (shapeNet) -- (outputfun);

\end{tikzpicture}
  \caption{Architecture of NsPOD.}
  \label{fig:nspod_scheme}
\end{figure}
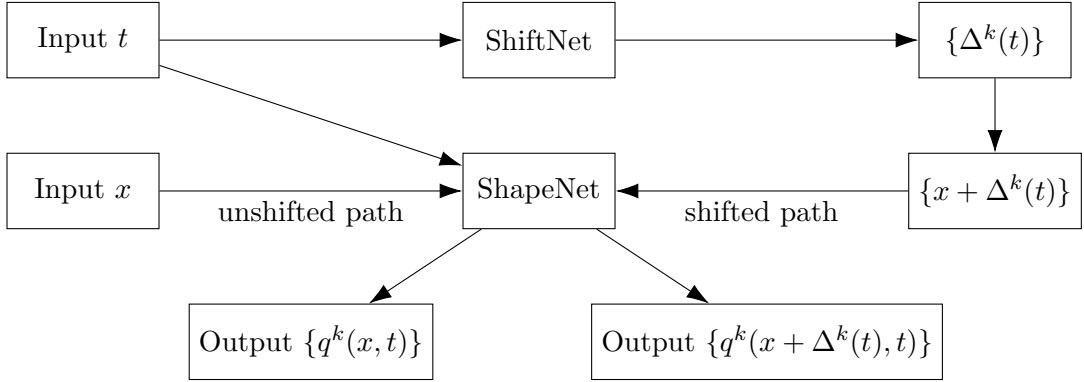

To solve~\eqref{eq:discr_pb} jointly, we propose a neural
network approach.
Making use of the translation symmetry of the underlying transport, we divide the network into two sub-networks: one of them is
dedicated to learning the shape of the co-moving fields $\{q^{k}\}$ and is called
ShapeNet, while the other one is estimating the transport shifts $\{\mT^{k}\}$ and is
entitled $\mathrm{ShiftNet}$.
In particular, ShiftNet is designed to estimate the shifts $\{\Delta^{k}\}$.
The overall architecture of the network is displayed in
Figure~\ref{fig:nspod_scheme} and we call our neural network sPOD approach: NsPOD\@.
NsPOD takes a time value $t$ and a spatial point $x$ as inputs.
The time value is fed into ShiftNet that computes the different shifts $\{\Delta^{k}(t)\}$
which are then used to shift the point $x$.
These shifted points $\{x+\Delta^{k}(t)\}$ are then given together with the inputs
$(x,t)$ to ShapeNet which outputs both $\{q^{k}(x,t)\}$ and $\{q^{k}(x+\Delta^{k}(t),t)\}$.
Notice that the outputs $\{q^{k}\}$ should have low rank while the sum of
the outputs $\{q^{k}(x+\Delta^{k}(t),t)\}$ should be close to $q$. 
Here, ShiftNet and ShapeNet are black box models for $\{q^{k}\}$ and $\{\Delta^{k}\}$ built
from the knowledge of the snapshot matrix $\*Q$ and the discretized grid on $\Omega\times\tau$.
In practice, ShapeNet and ShiftNet are built each from $K$ blocks, one for each $q^{k}$
and $\Delta^{k}$.

\paragraph{ShapeNet architecture.}
ShapeNet is a fully connected neural network composed of three linear hidden layers
interleaved with an Exponential Linear Unit (ELU) activation function.
On the other hand, the input layer has two neurons, one for the time point $t$ and one 
for the spatial point $x$, whereas the output layer is fully connected
with no activation function.


\paragraph{ShiftNet architecture.}
ShiftNet is first implemented as a simple polynomial regression model.
However, for more intricate problems such as the ones met in real-life 
applications, the polynomial regression is a too naive approach.
We replace it with a network similar to ShapeNet, i.e. composed of three fully
connected layers, each of them being followed by an ELU activation function.
This architecture offers superior generalization capabilities compared to the
simple polynomial regression models.
Indeed, it is less prone to overfitting and consequently, is able to capture 
complex dynamics more effectively without being overly tailored to the specific
traits of the initial training data.
Therefore, in Section~\ref{sec:simul}, we use the simple polynomial regression
model for the toy example and the fully connected network for the real-life
application.


\paragraph{Training.}
Because we want to solve~\eqref{eq:discr_pb} jointly, the two sub-networks
are connected and the training is performed simultaneously on both of
them.
The training is done by performing backpropagation with $\Loss$, the
objective function of~\eqref{eq:discr_pb}, as the loss function.
However, $\Loss$ is not differentiable at points where $\*Q^{k}$ has at least one
singular value equal to zero due to the nuclear norms in the regularization.
As a consequence, the gradient of $\Loss$ does not exist at these points.
To circumvent this issue, we use a subgradient of the nuclear norm at the points
where $\Loss$ is not differentiable.
Hence, if $\*Q$ is an $M \times N$ matrix of rank $R$ with singular value
decomposition $\*Q=\*U\bSigma\*V^\top$, then the subdifferential of the nuclear norm
at $\*Q$ is given by~\cite{Watson_G_1992_j-lin-algebra-appl_characterization_ssmn}
\begin{equation}
  \label{eq:diff_nucnorm}
  \partial \nucnorm{\*Q} =
  \{\*U\*V^{\top} + \*W \mid \spnorm{\*W} \leq 1, \, \*U^{\top}\*W  = \*0, \, \*W\*V = \*0\}
  \, .
\end{equation}
In other words, the subgradient of the nuclear norm at $\*W$ are the matrices
of the form $\*U\*V^{\top} + \*W$, where $\*W$ is an $M \times N$ matrix with spectral
norm lower than $1$ and whose column space is orthogonal to the column space of
$\*U$ and to the row space of $\*V$.
As shown in~\eqref{eq:diff_nucnorm}, the subdifferential is a set at the point where
the function is non-differentiable.
Consequently, several choices of subgradients are possible.
For the sake of simplicity, we choose the subgradient vector corresponding to
$\*W = \*0$, which is a common practice in the literature~\cite{Candes_E_2011_j-acm_robust_pca}.

The design of NsPOD, in particular its input layer, allows us to use each of $M\times N$ coordinates $(x, t)$ as input and the respective data points contained in $\*Q$ as targets.
We also note that for our current work, we do not use mini-batches for training rather we use the entire data set at once. 



\paragraph{Refinement.}
Once the network is trained, we can refine the estimation of ShapeNet by using
existing sPOD methods.
Indeed, we can use the results of the ShiftNet as the a priori knowledge of the
transformations $\{\mathcal{T}^{k}\}$ and the results of ShapeNet as the initial value for
$\{q^k\}$ in existing sPOD methods, such as sPOD-ALM 
from~\cite{Krah_P_2024_PP_robust_spodpmdfmt}, in order to perform a warm start
of these methods.
This additional step helps to improve the predicted co-moving fields $\{q^k\}$,
as shown in \cref{sec:simul}.
Furthermore, we note that, in the realm of time-parameter predictions, the NsPOD
needs no additional training to estimate unseen parameter values.
This is a clear advantage over the shifted POD-DL 
approach~\cite{Burela_S_2023_PP_parametrc_morwfmspbdlm}, which requires the
additional training of a neural network after decomposing the data with the sPOD. 


\section{Numerical results}
\label{sec:simul}

We demonstrate the efficiency of our approach on a synthetic test case and a
real-world scenario of a spreading wildland fire.
All the simulations presented in this section have been conducted on Python 3.9,
PyTorch 2.2.2, and CUDA 1108, and have been run in the mesocenter of
Aix-Marseille Universit{\'e} on an 8-core CPU shipped with 16GB of memory and a
Pascal GPU\@.
The training of NsPOD is performed with the Adaptive Moment Estimation (Adam) method ~\cite{Kingma_D_2014_PP_adam_mso}.


\paragraph{Stopping criterion.}
We use the following stopping criterion for NsPOD, which is based on
the relative decrease in the loss function between two epochs $e$ and $e+1$:
\begin{equation}
    \Loss(\*x^{(e)}) - \Loss(\*x^{(e+1)}) \leq \delta \Loss(\*x^{(e)}) \, ,
\end{equation}
where $\delta$ is a
tolerance set to $10^{-4}$ while $\*x^{(e)}$ and $\*x^{(e+1)}$ represent the
vector of outputs of the model, which is $(\{q^{k}\},\{\mT^{k}\})$, at two consecutive
epochs.
When the criterion is violated or when the number of epochs exceeds the maximum
limit, the training process is complete.


\paragraph{Performance evaluation.}
We measure the performance of our approach with two metrics: the relative
reconstruction error between the snapshot matrix and the computed decomposition
\begin{equation*}
  E^{\mathrm{NN}}_{\mathrm{rec}} = \frac{\fnorm{\*Q - \sum_{k=1}^{K}\*T^{k}\*Q^{k}}}{\fnorm{\*Q}}
  \, , 
\end{equation*}
and the rank of each co-moving field.
In particular, Table~\ref{tab:error} shows the reconstruction error for our two test cases.
\begin{table}[!t]
  \caption{Performance analysis of NsPOD for the test cases.}
  \begin{center}
    \setlength\tabcolsep{13pt}
    \begin{tabular}{lllll}
    \toprule
    \multirow{2}{*}{}                          & \multicolumn{1}{c}{NsPOD}                 & \multicolumn{3}{c}{sPOD-ALM} \\
                                                        & $E^{\mathrm{NN}}_{\mathrm{rec}}$          & ranks           & $E_{\mathrm{rec}}$  & $N_\mathrm{iter}$   \\
    \midrule
    \multirow{1}{*}{Crossing waves}                     & \multirow{2}{*}{3.19e-02} & (1, 1)         & 9.24e-02   & 14      \\
                                                        &                       & (2, 2)      & 6.36e-03   &    60\\
    \midrule
    \multirow{5}{*}{Wildland fire}                      & \multirow{4}{*}{2.08e-02} & (1, 1, 1)      & 1.58e-02      & 36\\
                                                        &                       & (2, 1, 2)      & 1.06e-02      & 49\\
                                                        &                       & (3, 2, 3)      & 2.61e-03      & 268\\
                                                        &                       & (4, 4, 4)      & 1.74e-03     & 951\\
    \bottomrule
    \end{tabular}
  \end{center}
\label{tab:error}
\end{table}


\paragraph{Implementation.}
The choice of the regularization parameter $\lambda$ is performed empirically:
we have tested several values and we have chosen the one that yields the best results.
The initialization of the parameters of the ShapeNet and ShiftNet are performed
randomly using a uniform distribution $\mU(-1/\sqrt{2},1/\sqrt{2})$.
The implementation parameters for NsPOD are listed in Table~\ref{tab:NN_params} with $N_\mathrm{epochs}$ being the prescribed maximum number of epochs and $\alpha$
being the learning rate.
In addition, the seeds used for random initialization of the weights of both
sub-networks are given for reproducibility purposes.

\begin{table}[htp!]
  \begin{center}
    \begin{minipage}{\textwidth}
      \caption{NsPOD parameters for the test cases}
      \label{tab:NN_params}
      \centering
      \begin{tabular}{l c c c c c}
        \toprule
        Model & $N_\mathrm{epochs}$ & $\alpha$ & $\lambda$ & seed & training time\\
        \midrule
        Crossing waves & $10^6$ & $0.001$ & $0.05$ & $54$ & $17h$\\ 
        Wildland-fire & $10^5$ & $0.001$ & $0.1$ & $420$ & $4h$\\
        \bottomrule
      \end{tabular}
    \end{minipage}
  \end{center}
\end{table}


\subsection{Synthetic data case (Crossing waves)}
\label{ssec:simu_toy}

We start by testing NsPOD on synthetic data to investigate the behavior of
our method and the sensitivity to the initialization of the weights.
We build the transport-dominated field $q$ from the superposition of two Gaussian
traveling waves,
\begin{equation}
  \label{eq:crosslines}
  q(x, t) = \mathrm{sin}\left(\frac{\pi t}{10}\right)\mathrm{exp}\left(-\frac{{\left(x - \mu - \Delta^1(t)\right)}^2}{\sigma^2}\right)
  + \mathrm{cos}\left(\frac{\pi t}{10}\right)\mathrm{exp}\left(-\frac{{\left(x - \mu - \Delta^2(t)\right)}^2}{\sigma^2}\right)
  \, ,
\end{equation}
where $(\mu,\sigma)=(200,4)$.
The two shifts $\Delta^{1}$ and $\Delta^{2}$ define the transport $\mT^{1}$ and $\mT^{2}$ and are
given by
\begin{equation*}
  \Delta^1(t) = 0.15t^3 + 0.8t + 1.5
  \quad \text{and} \quad
  \Delta^2(t) = -18t + 2 \, .
\end{equation*}
The spatial and temporal domains are given by $\Omega=[0, 400]$ and $\tau=[-10, 10]$,
respectively.
They are discretized with $M=400$ equally spaced grid points and $N=200$ temporal grid
points that are uniformly distributed resulting in a snapshot matrix $\*Q\in\RR^{400\times 200}$.

For this example, we use a polynomial regression for ShiftNet since we
have arbitrarily chosen polynomial functions for the definition of the shifts
\begin{equation}
  \mathrm{ShiftNet}^1(t) = w_{31} t^3 + w_{21} t^2 + w_{11} t + w_{01}  \quad \text{and} \quad
  \mathrm{ShiftNet}^2(t) = w_{12} t + w_{02} \, ,
\end{equation}
where the neural network framework learns the correct coefficients
$w_{ij}$ and the corresponding representation of the co-moving frames. 

\begin{figure}[htp!]
  \centering
  \begin{subfigure}{0.9\textwidth}
  \setlength\figureheight{0.31\linewidth}
  \setlength\figurewidth{0.25\linewidth}
\begin{tikzpicture}

\begin{groupplot}[group style={group size=6 by 1, horizontal sep=0.2cm}]
\nextgroupplot[
height=\figureheight,
tick align=outside,
tick pos=left,
title={\(\displaystyle \mathbf{Q}\)},
width=\figurewidth,
xlabel={\(\displaystyle x\)},
x grid style={darkgray176},
xmin=0, xmax=200,
xtick=\empty,
ytick=\empty,
xtick style={color=black},
ylabel={\(\displaystyle t\)},
y grid style={darkgray176},
ymin=0, ymax=400,
axis line style={draw=none},
tick style={draw=none},
ytick style={color=black},
]
\addplot graphics [includegraphics cmd=\pgfimage,xmin=0, xmax=200, ymin=0, ymax=400] {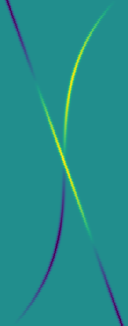};

\nextgroupplot[
height=\figureheight,
tick align=outside,
tick pos=left,
title={\(\displaystyle \tilde{\mathbf{Q}}\)},
width=\figurewidth,
xlabel={\(\displaystyle x\)},
x grid style={darkgray176},
xmin=0, xmax=200,
xtick=\empty,
ytick=\empty,
xtick style={color=black},
y grid style={darkgray176},
ymin=0, ymax=400,
axis line style={draw=none},
tick style={draw=none},
ytick style={color=black},
]
\addplot graphics [includegraphics cmd=\pgfimage,xmin=0, xmax=200, ymin=0, ymax=400] {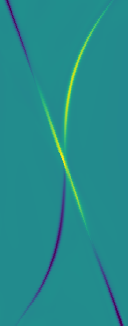};

\nextgroupplot[
height=\figureheight,
tick align=outside,
tick pos=left,
title={\(\displaystyle \*T^1\mathbf{Q}^1\)},
width=\figurewidth,
xlabel={\(\displaystyle x\)},
x grid style={darkgray176},
xmin=0, xmax=200,
xtick=\empty,
ytick=\empty,
xtick style={color=black},
y grid style={darkgray176},
ymin=0, ymax=400,
axis line style={draw=none},
tick style={draw=none},
ytick style={color=black},
]
\addplot graphics [includegraphics cmd=\pgfimage,xmin=0, xmax=200, ymin=0, ymax=400] {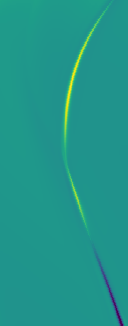};

\nextgroupplot[
height=\figureheight,
tick align=outside,
tick pos=left,
title={\(\displaystyle \*T^2\mathbf{Q}^2\)},
width=\figurewidth,
xlabel={\(\displaystyle x\)},
x grid style={darkgray176},
xmin=0, xmax=200,
xtick=\empty,
ytick=\empty,
xtick style={color=black},
y grid style={darkgray176},
ymin=0, ymax=400,
axis line style={draw=none},
tick style={draw=none},
ytick style={color=black},
]
\addplot graphics [includegraphics cmd=\pgfimage,xmin=0, xmax=200, ymin=0, ymax=400] {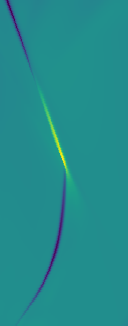};

\nextgroupplot[
height=\figureheight,
tick align=outside,
tick pos=left,
title={\(\displaystyle \mathbf{Q}^1\)},
width=\figurewidth,
xlabel={\(\displaystyle x\)},
x grid style={darkgray176},
xmin=0, xmax=200,
xtick=\empty,
ytick=\empty,
xtick style={color=black},
y grid style={darkgray176},
ymin=0, ymax=400,
axis line style={draw=none},
tick style={draw=none},
ytick style={color=black},
]
\addplot graphics [includegraphics cmd=\pgfimage,xmin=0, xmax=200, ymin=0, ymax=400] {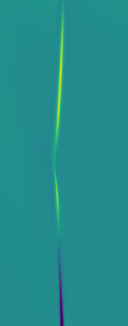};

\nextgroupplot[
    colorbar,
    colorbar style={ytick={-1.00, -0.50, 0, 0.50, 1.00},yticklabels={
      \(\displaystyle {\ensuremath{-}1.00}\),
      \(\displaystyle {\ensuremath{-}0.50}\),
      \(\displaystyle {0.0}\),
      \(\displaystyle {0.50}\),
      \(\displaystyle {1.00}\)
    },ylabel={}},
    colormap/viridis,
    height=\figureheight,
    point meta max=1.05024907387885,
    point meta min=-1.05,
tick align=outside,
tick pos=left,
colorbar/width=2mm,
title={\(\displaystyle \mathbf{Q}^2\)},
width=\figurewidth,
xlabel={\(\displaystyle x\)},
x grid style={darkgray176},
xmin=0, xmax=200,
xtick=\empty,
ytick=\empty,
xtick style={color=black},
y grid style={darkgray176},
ymin=0, ymax=400,
axis line style={draw=none},
tick style={draw=none},
ytick style={color=black},
]
\addplot graphics [includegraphics cmd=\pgfimage,xmin=0, xmax=200, ymin=0, ymax=400] {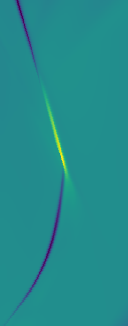};
\end{groupplot}

\end{tikzpicture}
    \vspace{-0.5cm}
    \caption{Initialization resulting in undesirable separation of the transports.}
    \label{fig:crosslines_seed7}
  \end{subfigure}
  \begin{subfigure}{0.9\textwidth}
  \setlength\figureheight{0.31\linewidth}
  \setlength\figurewidth{0.25\linewidth}
\begin{tikzpicture}

\begin{groupplot}[group style={group size=6 by 1, horizontal sep=0.2cm}]
\nextgroupplot[
    height=\figureheight,
    tick align=outside,
    tick pos=left,
    title={\(\displaystyle \mathbf{Q}\)},
    width=\figurewidth,
    xlabel={\(\displaystyle x\)},
    x grid style={darkgray176},
    xmin=0, xmax=200,
    xtick=\empty,
    ytick=\empty,
    xtick style={color=black},
    ylabel={\(\displaystyle t\)},
    y grid style={darkgray176},
    ymin=0, ymax=400,
    axis line style={draw=none},
    tick style={draw=none},
ytick style={color=black},
]
\addplot graphics [includegraphics cmd=\pgfimage,xmin=0, xmax=200, ymin=0, ymax=400] {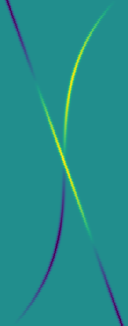};

\nextgroupplot[
    height=\figureheight,
    tick align=outside,
    tick pos=left,
    title={\(\displaystyle \tilde{\mathbf{Q}}\)},
    width=\figurewidth,
    xlabel={\(\displaystyle x\)},
    x grid style={darkgray176},
    xmin=0, xmax=200,
    xtick=\empty,
    ytick=\empty,
    xtick style={color=black},
    y grid style={darkgray176},
    ymin=0, ymax=400,
    axis line style={draw=none},
    tick style={draw=none},
    ytick style={color=black},
]
\addplot graphics [includegraphics cmd=\pgfimage,xmin=0, xmax=200, ymin=0, ymax=400] {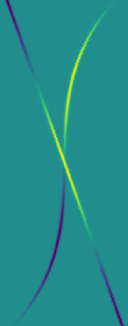};

\nextgroupplot[
    height=\figureheight,
    tick align=outside,
    tick pos=left,
    title={$\displaystyle \*T^{1}\*Q^{1}$},
    width=\figurewidth,
    xlabel={$\displaystyle x$},
    x grid style={darkgray176},
    xmin=0, xmax=200,
    xtick=\empty,
    ytick=\empty,
    xtick style={color=black},
    y grid style={darkgray176},
    ymin=0, ymax=400,
    axis line style={draw=none},
    tick style={draw=none},
    ytick style={color=black},
]
\addplot graphics [includegraphics cmd=\pgfimage,xmin=0, xmax=200, ymin=0, ymax=400] {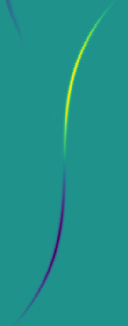};

\nextgroupplot[
    height=\figureheight,
    tick align=outside,
    tick pos=left,
    title={$\displaystyle \*T^{2}\*Q^{2}$},
    width=\figurewidth,
    xlabel={$\displaystyle x$},
    x grid style={darkgray176},
    xmin=0, xmax=200,
    xtick=\empty,
    ytick=\empty,
    xtick style={color=black},
    y grid style={darkgray176},
    ymin=0, ymax=400,
    axis line style={draw=none},
    tick style={draw=none},
    ytick style={color=black},
]
\addplot graphics [includegraphics cmd=\pgfimage,xmin=0, xmax=200, ymin=0, ymax=400] {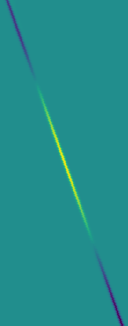};

\nextgroupplot[
    height=\figureheight,
    tick align=outside,
    tick pos=left,
    title={\(\displaystyle \mathbf{Q}^1\)},
    width=\figurewidth,
    xlabel={\(\displaystyle x\)},
    x grid style={darkgray176},
    xmin=0, xmax=200,
    xtick=\empty,
    ytick=\empty,
    xtick style={color=black},
    y grid style={darkgray176},
    ymin=0, ymax=400,
    axis line style={draw=none},
    tick style={draw=none},
    ytick style={color=black},
]
\addplot graphics [includegraphics cmd=\pgfimage,xmin=0, xmax=200, ymin=0, ymax=400] {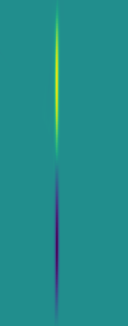};

\nextgroupplot[
    colorbar,
    colorbar style={ytick={-1.00, -0.50, 0, 0.50, 1.00},yticklabels={
      \(\displaystyle {\ensuremath{-}1.00}\),
      \(\displaystyle {\ensuremath{-}0.50}\),
      \(\displaystyle {0.0}\),
      \(\displaystyle {0.50}\),
      \(\displaystyle {1.00}\)
    },ylabel={}},
    colormap/viridis,
    height=\figureheight,
    point meta max=1.05024907387885,
    point meta min=-1.05,
    tick align=outside,
    tick pos=left,
    colorbar/width=2mm,
    title={\(\displaystyle \mathbf{Q}^2\)},
    width=\figurewidth,
    xlabel={\(\displaystyle x\)},
    x grid style={darkgray176},
    xmin=0, xmax=200,
    xtick=\empty,
    ytick=\empty,
    xtick style={color=black},
    y grid style={darkgray176},
    ymin=0, ymax=400,
    axis line style={draw=none},
    tick style={draw=none},
    ytick style={color=black},
]
\addplot graphics [includegraphics cmd=\pgfimage,xmin=0, xmax=200, ymin=0, ymax=400] {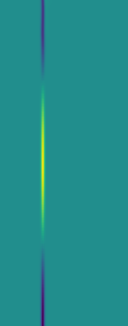};
\end{groupplot}

\end{tikzpicture}
    \vspace{-0.5cm}
    \caption{Initialization resulting in desirable separation of the transports.}
    \label{fig:crosslines_seed1}
  \end{subfigure}
    \begin{subfigure}{0.9\textwidth}
  \setlength\figureheight{0.31\linewidth}
  \setlength\figurewidth{0.25\linewidth}
\begin{tikzpicture}

\begin{groupplot}[group style={group size=6 by 1, horizontal sep=0.2cm}]
\nextgroupplot[
    height=\figureheight,
    tick align=outside,
    tick pos=left,
    title={\(\displaystyle \mathbf{Q}\)},
    width=\figurewidth,
    xlabel={\(\displaystyle x\)},
    x grid style={darkgray176},
    xmin=0, xmax=200,
    xtick=\empty,
    ytick=\empty,
    xtick style={color=black},
    ylabel={\(\displaystyle t\)},
    y grid style={darkgray176},
    ymin=0, ymax=400,
    axis line style={draw=none},
    tick style={draw=none},
ytick style={color=black},
]
\addplot graphics [includegraphics cmd=\pgfimage,xmin=0, xmax=200, ymin=0, ymax=400] {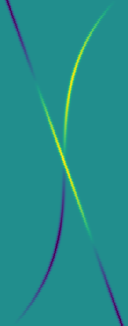};

\nextgroupplot[
    height=\figureheight,
    tick align=outside,
    tick pos=left,
    title={\(\displaystyle \tilde{\mathbf{Q}}\)},
    width=\figurewidth,
    xlabel={\(\displaystyle x\)},
    x grid style={darkgray176},
    xmin=0, xmax=200,
    xtick=\empty,
    ytick=\empty,
    xtick style={color=black},
    y grid style={darkgray176},
    ymin=0, ymax=400,
    axis line style={draw=none},
    tick style={draw=none},
    ytick style={color=black},
]
\addplot graphics [includegraphics cmd=\pgfimage,xmin=0, xmax=200, ymin=0, ymax=400] {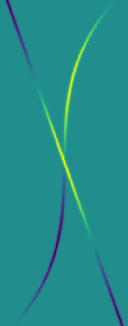};

\nextgroupplot[
    height=\figureheight,
    tick align=outside,
    tick pos=left,
    title={$\displaystyle \*T^{1}\*Q^{1}$},
    width=\figurewidth,
    xlabel={$\displaystyle x$},
    x grid style={darkgray176},
    xmin=0, xmax=200,
    xtick=\empty,
    ytick=\empty,
    xtick style={color=black},
    y grid style={darkgray176},
    ymin=0, ymax=400,
    axis line style={draw=none},
    tick style={draw=none},
    ytick style={color=black},
]
\addplot graphics [includegraphics cmd=\pgfimage,xmin=0, xmax=200, ymin=0, ymax=400] {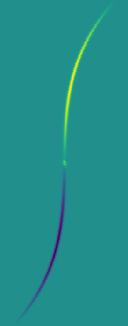};

\nextgroupplot[
    height=\figureheight,
    tick align=outside,
    tick pos=left,
    title={$\displaystyle \*T^{2}\*Q^{2}$},
    width=\figurewidth,
    xlabel={$\displaystyle x$},
    x grid style={darkgray176},
    xmin=0, xmax=200,
    xtick=\empty,
    ytick=\empty,
    xtick style={color=black},
    y grid style={darkgray176},
    ymin=0, ymax=400,
    axis line style={draw=none},
    tick style={draw=none},
    ytick style={color=black},
]
\addplot graphics [includegraphics cmd=\pgfimage,xmin=0, xmax=200, ymin=0, ymax=400] {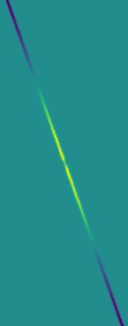};

\nextgroupplot[
    height=\figureheight,
    tick align=outside,
    tick pos=left,
    title={\(\displaystyle \mathbf{Q}^1\)},
    width=\figurewidth,
    xlabel={\(\displaystyle x\)},
    x grid style={darkgray176},
    xmin=0, xmax=200,
    xtick=\empty,
    ytick=\empty,
    xtick style={color=black},
    y grid style={darkgray176},
    ymin=0, ymax=400,
    axis line style={draw=none},
    tick style={draw=none},
    ytick style={color=black},
]
\addplot graphics [includegraphics cmd=\pgfimage,xmin=0, xmax=200, ymin=0, ymax=400] {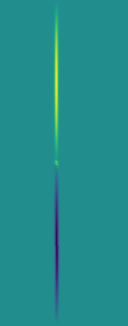};

\nextgroupplot[
    colorbar,
    colorbar style={ytick={-1.00, -0.50, 0, 0.50, 1.00},yticklabels={
      \(\displaystyle {\ensuremath{-}1.00}\),
      \(\displaystyle {\ensuremath{-}0.50}\),
      \(\displaystyle {0.0}\),
      \(\displaystyle {0.50}\),
      \(\displaystyle {1.00}\)
    },ylabel={}},
    colormap/viridis,
    height=\figureheight,
    point meta max=1.05024907387885,
    point meta min=-1.05,
    tick align=outside,
    tick pos=left,
    colorbar/width=2mm,
    title={\(\displaystyle \mathbf{Q}^2\)},
    width=\figurewidth,
    xlabel={\(\displaystyle x\)},
    x grid style={darkgray176},
    xmin=0, xmax=200,
    xtick=\empty,
    ytick=\empty,
    xtick style={color=black},
    y grid style={darkgray176},
    ymin=0, ymax=400,
    axis line style={draw=none},
    tick style={draw=none},
    ytick style={color=black},
]
\addplot graphics [includegraphics cmd=\pgfimage,xmin=0, xmax=200, ymin=0, ymax=400] {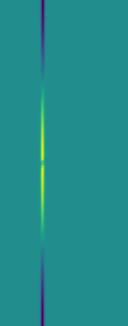};
\end{groupplot}

\end{tikzpicture}
    \vspace{-0.05cm}
    \caption{Results after cleaning with sPOD-ALM and initialization with NsPOD.}
    \label{fig:crosslines_sPOD}
  \end{subfigure}
  \caption{Results of NsPOD on crossing waves test case for two different initialization seeds (a) and (b). Furthermore, (c) shows the results of the sPOD-ALM using the shifts and co-moving fields determined by NsPOD.}
  \label{fig:crosslines_ex}
\end{figure}

Figure~\ref{fig:crosslines_ex} shows the results of NsPOD for the decomposition
of $q$.
In Figure~\ref{fig:crosslines_seed1}, we observe that the two crossing lines are
well separated by NsPOD, albeit a minor disturbance in $\*T^1 \*Q^1$ plot, and that
the predicted co-moving fields $\*Q^{1}$ and $\*Q^{2}$ have low rank as well.
Furthermore, we see in Table~\ref{tab:error} that NsPOD also performs well in
terms of reconstruction error measure.
When the results from NsPOD are used to perform a warm start of sPOD-ALM, the minor
artifact left in $\*T^1 \*Q^1$ disappears and we observe a clean separation as
shown in Figure~\ref{fig:crosslines_sPOD}.

However, due to the non-uniqueness of the sPOD ansatz~\cite{Krah_P_2024_PP_robust_spodpmdfmt}, the initialization of the weights in NsPOD networks is crucial for achieving the desired decomposition.
This is demonstrated in~\cref{fig:crosslines_seed7}, where improper initialization
results in an undesirable decomposition: the method does not recognize the continuity
of the waves after the crossing point.
Thus, we note that the initialization sensitively impacts the performance and the
separation capacity of NsPOD.
In this test case, 17 different seed values were tested. Among them, 6 seeds led to an appropriate
decomposition, with low-rank co-moving fields and accurate shifts.
\begin{figure}[htp!]
\centering
  \setlength\figureheight{0.32\linewidth}
  \setlength\figurewidth{0.2\linewidth}
\begin{tikzpicture}

\begin{groupplot}[group style={group size=5 by 1, horizontal sep=0.4cm}]
\nextgroupplot[
height=\figureheight,
tick align=outside,
tick pos=left,
title={\(\displaystyle \mathbf{Q}\)},
width=1.3*\figurewidth,
xlabel={\(\displaystyle t\)},
x grid style={darkgray176},
xmin=0, xmax=200,
xtick=\empty,
ytick=\empty,
xtick style={color=black},
ylabel={\(\displaystyle x\)},
y grid style={darkgray176},
ymin=0, ymax=400,
axis line style={draw=none},
tick style={draw=none},
ytick style={color=black},
]
\addplot graphics [includegraphics cmd=\pgfimage,xmin=0, xmax=200, ymin=0, ymax=400] {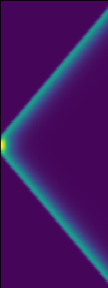};

\nextgroupplot[
height=\figureheight,
tick align=outside,
tick pos=left,
title={\(\displaystyle \tilde{\mathbf{Q}}\)},
width=1.3*\figurewidth,
xlabel={\(\displaystyle t\)},
x grid style={darkgray176},
xmin=0, xmax=200,
xtick=\empty,
ytick=\empty,
xtick style={color=black},
y grid style={darkgray176},
ymin=0, ymax=400,
axis line style={draw=none},
tick style={draw=none},
ytick style={color=black},
]
\addplot graphics [includegraphics cmd=\pgfimage,xmin=0, xmax=200, ymin=0, ymax=400] {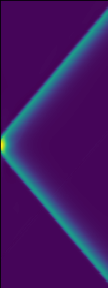};

\nextgroupplot[
height=\figureheight,
tick align=outside,
tick pos=left,
title={\(\displaystyle \*T^1\mathbf{Q}^1\)},
width=1.3*\figurewidth,
xlabel={\(\displaystyle t\)},
x grid style={darkgray176},
xmin=0, xmax=200,
xtick=\empty,
ytick=\empty,
xtick style={color=black},
y grid style={darkgray176},
ymin=0, ymax=400,
axis line style={draw=none},
tick style={draw=none},
ytick style={color=black},
]
\addplot graphics [includegraphics cmd=\pgfimage,xmin=0, xmax=200, ymin=0, ymax=400] {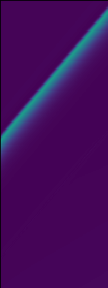};

\nextgroupplot[
height=\figureheight,
tick align=outside,
tick pos=left,
title={\(\displaystyle \*T^2\mathbf{Q}^2\)},
width=1.3*\figurewidth,
xlabel={\(\displaystyle t\)},
x grid style={darkgray176},
xmin=0, xmax=200,
xtick=\empty,
ytick=\empty,
xtick style={color=black},
y grid style={darkgray176},
ymin=0, ymax=400,
axis line style={draw=none},
tick style={draw=none},
ytick style={color=black},
]
\addplot graphics [includegraphics cmd=\pgfimage,xmin=0, xmax=200, ymin=0, ymax=400] {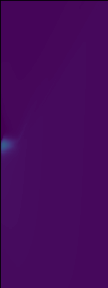};

\nextgroupplot[
colorbar,
colorbar style={ytick={0,0.2,0.4,0.6,0.8,1},yticklabels={
  \(\displaystyle {0.0}\),
  \(\displaystyle {0.2}\),
  \(\displaystyle {0.4}\),
  \(\displaystyle {0.6}\),
  \(\displaystyle {0.8}\),
  \(\displaystyle {1.0}\)
},ylabel={}, width=7},
colormap/viridis,
height=\figureheight,
tick align=outside,
tick pos=left,
title={\(\displaystyle \*T^3\mathbf{Q}^3\)},
width=1.3*\figurewidth,
xlabel={\(\displaystyle t\)},
x grid style={darkgray176},
xmin=0, xmax=200,
xtick=\empty,
ytick=\empty,
xtick style={color=black},
y grid style={darkgray176},
ymin=0, ymax=400,
axis line style={draw=none},
tick style={draw=none},
ytick style={color=black},
]
\addplot graphics [includegraphics cmd=\pgfimage,xmin=0, xmax=200, ymin=0, ymax=400] {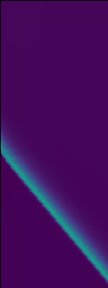};

\end{groupplot}

\end{tikzpicture}  
  \caption{Temperature field of the wildland-fire model test case decomposed with NsPOD.}
  \label{fig:wildland_fire1D_nn}
\end{figure}





\subsection{Wildland-fire model}

We now test the NsPOD on the 1D wildland fire model taken
from~\cite{Burela_S_2023_PP_parametrc_morwfmspbdlm}.
The wildland fire model is a coupled non-linear reaction-diffusion equation that
describes the evolution of the temperature and the fuel supply mass fraction
with time.
The differential equations are discretized using $M = 250$ equally spaced grid
points on $\Omega=[0,250\: \mathrm{m}]$ and solved up to time
$T_{\text{end}}=1400\:\mathrm{s}$ with $N =300$ time steps for a reaction rate
$\mu = 558.49\:\mathrm{K}$ and a wind velocity $v=0$m/s.
We only illustrate our method on the temperature but similar results can be
obtained on the supply mass fraction. 

In contrast with the previous example, we now use fully connected networks of
ShiftNet instead of a polynomial regression.
To obtain a good initialization of the network and to speed up the training, we
use a set of pre-trained weights specifically developed for scenario exhibiting
dynamic transport patterns, similar to those found in wildland fires.
The pre-trained weights are derived from a model trained on synthetic data
constructed to simulate two co-moving frames within a spatial domain $\Omega:=[0, 400]$ and time domain $\tau:=[0, 200]$.
The synthetic data model is defined as
\begin{equation*}
  q(x, t) = f(x - \mu_1 - \Delta^1(t)) + f(x - \mu_2 - \Delta^2(t)) \, ,
\end{equation*}
where $f(x)=\exp(-{x}^{2}/\delta^{2})$, $\delta=4.0$, $\mu_1=110$, $\mu_2=200$, $\Delta^1(t)=-7t+1$ and
$\Delta^2(t)=6t-1$.
For the ignition component of the model, we continue to use a random
initialization but we ensure consistency by setting a specific seed, as done in
previous examples, to guarantee the reproducibility of our results.
Using a pre-trained set of weights offers a strong starting point for the model,
allowing to effectively capture the essential dynamics of fire propagation
right from the beginning.
We note that appropriate pre-training is necessary for efficiency and is a general practice in the field of image recognition.
The advantage of our approach is that the pre-training can be done on a smaller data set since our optimization problem is formulated in a continuous setting.

Figure~\ref{fig:wildland_fire1D_nn} displays the results of NsPOD applied to the
1D wildland-fire model.
We observe a successful separation of the temperature field into three co-moving
frames, as shown in \cref{fig:wildland_fire1D_nn}.
The fields $\*Q^{1}$ and $\*Q^3$ have rank $2$ while the field $\*Q^{2}$ corresponding to the ignition has rank $1$.
The corresponding reconstruction error measure is presented in Table~\ref{tab:error}.


\section{Conclusion}
\label{sec:conclusion}

We have considered the problem of the joint estimation of the transports and the
co-moving fields in the sPOD.
In contrast with previous works, our approach has started with the continuous
and joint formulation of the sPOD problem.
We have then adopted a neural network approach with a twin architecture to tackle
this optimization problem and we have leveraged existing methods to improve the outputs of the
model.
Although our approach is sensitive to the initialization of the network, it has
shown great results in separating the different components of the dynamics on a
wildland fire model.
To stabilize the network, a promising future idea is to add physics information to our
network architecture.


\section*{Code availability}
The code and a Jupyter notebook presenting our results is publicly available:
\begin{center}
\urlstyle{tt}
\url{https://github.com/MOR-transport/automated_NsPOD}
\end{center}

\section*{Credit authorship contribution statement}

\noindent{}
In the following, we declare the authors' contributions to this work. \\
{\small
\noindent
\begin{tabular}{@{}p{0.17\linewidth} p{0.77\linewidth}}
\textbf{B. Zorawski:} & Software, Visualization, Writing – review \& editing \\
\textbf{S. Burela} & Software, Visualization, Writing – review \& editing\\
\textbf{P. Krah:} & Supervision, Methodology, Initial Idea, Writing - original draft.\\
\textbf{A. Marmin:} & Supervision, Methodology, Writing – original draft.\\
\textbf{K. Schneider:} & Supervision, Funding Acquisition,  Writing – review \& editing.
\end{tabular}
}


\section*{Acknowledgment}
{\small
P. Krah is supported by ANR-20-CE46-0010, CM2E. S. Burela acknowledges support from the DFG, GRK2433 DAEDALUS. B. Zorawski thanks DAAD for ERASMUS+ funding in Marseille.
The authors accessed HPC resources of IDRIS under Allocation A0142A14152, project No. 2023-91664 by GENCI, and Centre de Calcul Intensif d’Aix-Marseille Université.
}


\bibliographystyle{myplain}
\bibliography{abbr,refs}

\end{document}